# UPMAD-Net: A Brain Tumor Segmentation Network with Uncertainty Guidance and Adaptive Multimodal Feature Fusion

Zhanyuan Jia[1] ·Ni Yao[1]·Danyang Sun[1]·Chuang Han[1]·Yanting Li[1]·Jiaofen Nan[1]·Fubao Zhu[1,*] ·Chen Zhao[2*]·Weihua Zhou[3,4]

[1] School of Computer Science and Technology, Zhengzhou University of Light Industry, Zhengzhou 450002, Henan, China
[2] Department of Computer Science, Kennesaw State University Marietta, GA, USA
[3] Department of Applied Computing, Michigan Technological University, Houghton, MI, USA
[4] Center for Biocomputing and Digital Health, Institute of Computing and Cybersystems, and Health Research Institute, Michigan Technological University, Houghton, MI, USA

Zhanyuan Jia and Chen Zhao contributed equally.

[*] Correspondence:
Fubao Zhu
Email address: fbzhu@zzuli.edu.cn
Mailing address: School of Computer Science and Technology, Zhengzhou University of Light Industry, Zhengzhou 450002, Henan, China

Chen Zhao
Email address: czhao4@kennesaw.edu
Mailing address: 680 Arntson Dr, Marietta, GA 30060

**Abstract**

**Background:** Brain tumor segmentation has a significant impact on the diagnosis and treatment of brain tumors. Accurate brain tumor segmentation remains challenging due to their irregular shapes, vague boundaries, and high variability.

**Objective**: We propose a brain tumor segmentation method that combines deep learning with prior knowledge derived from a region-growing algorithm.

**Methods:** The proposed method utilizes a multi-scale feature fusion (MSFF) module and adaptive attention mechanisms (AAM) to extract multi-scale features and capture global contextual information. To enhance the model's robustness in low-confidence regions, the Monte Carlo Dropout (MC Dropout) strategy is employed for uncertainty estimation.

**Results**: Extensive experiments demonstrate that the proposed method achieves superior performance on Brain Tumor Segmentation (BraTS) datasets, significantly outperforming various state-of-the-art methods. On the BraTS2021 dataset, the test Dice scores are 89.18% for Enhancing Tumor (ET) segmentation, 93.67% for Whole Tumor (WT) segmentation, and 91.23% for Tumor Core (TC) segmentation. On the BraTS2019 validation set, the validation Dice scores are 87.43%, 90.92%, and 90.40% for ET, WT, and TC segmentation, respectively.

Ablation studies further confirmed the contribution of each module to segmentation accuracy, indicating that each component played a vital role in overall performance improvement.

**Conclusion**: This study proposed a novel 3D brain tumor segmentation network based on the U-Net architecture. By incorporating the prior knowledge and employing the uncertainty estimation method, the robustness and performance were improved.

The code for the proposed method is available at https://github.com/chenzhao2023/UPMAD_Net_BrainSeg.

## 1. Introduction

Brain tumors are serious disorders of the central nervous system, significantly affecting patients survival rates and quality of life[1,2]. Accurate segmentation of brain tumors from Magnetic resonance imaging (MRI) scans plays a vital role in clinical diagnosis, treatment planning, and prognosis assessment[3]. However, manual annotation by radiologists is time-consuming, subjective, and prone to inter-observer variability, which limits its practicality in large-scale clinical applications[4]. Consequently, automated segmentation methods have gained increasing attention for their potential to enhance both efficiency and consistency.

MRI, owing to its superior soft tissue contrast, has become the primary imaging modality for brain tumor analysis[5,6]. Common modalities—T1-weighted (T1), contrast-enhanced T1 (T1ce), Fluid-attenuated inversion recovery (FLAIR), and T2—provide (T2) complementary perspectives on tumor structure[7]. Multimodal MRI, by combining information across these sequences, offers a more comprehensive view of tumor characteristics, thereby improving segmentation accuracy[8]. However, tumor heterogeneity and cross-modality inconsistencies still pose significant challenges to accurate automatic segmentation in clinical practice.

Early approaches to brain tumor segmentation employed traditional image processing techniques such as thresholding, region growing, and edge detection[9]. Although intuitive and computationally efficient, these methods are highly sensitive to noise and fail to address the complex and heterogeneous nature of brain tumors[10]. Deep learning methods, especially convolutional neural networks (CNNs) and U-Net variants, have become dominant due to their ability to learn hierarchical feature representations[11,12]. However, their inherent locality bias and limited receptive field restrict their ability to model long-range dependencies, which are essential for segmenting tumors with irregular shapes or diffuse boundaries. To address these limitations, Transformer-based models such as TransBTS[13] and Swin UNETR[14] have been introduced, leveraging self-attention mechanisms to capture global context and improve semantic coherence[15]. Despite their success, they still face challenges in handling fine-grained boundaries and require substantial training data and computational resources, particularly in small or low-contrast tumor regions.

Despite the progress brought by Transformer-based architectures, brain tumor segmentation still faces inherent challenges due to the ambiguity and variability of tumor boundaries[16]. In MRI scans, transitions between tumor and healthy tissue are often gradual and indistinct, making it difficult to generate precise predictions, particularly in low-contrast or infiltrative regions. Traditional voxel-wise classification methods tend to produce overconfident outputs in these ambiguous areas, increasing the risk of misclassification. To address this issue, uncertainty estimation techniques have been increasingly incorporated into segmentation frameworks to enhance the reliability and interpretability of predictions[17,18]. By quantifying prediction confidence, these methods allow models to highlight regions of high uncertainty, which is particularly beneficial for guiding clinical decision-making and avoiding overconfident errors near uncertain boundaries[19]. For example, the Probabilistic U-Net proposed by Kohl et al.[20] generates multiple plausible segmentation hypotheses instead of a single deterministic mask, thereby capturing the inherent variability in ambiguous regions. Extending this idea, PHISeg[21] employs a hierarchical latent representation to further improve flexibility in modeling uncertainty. More recently, Chen et al.[22] integrated uncertainty-guided attention mechanisms into a Transformer-based architecture for brain tumor segmentation. This approach enables the model to focus more effectively on regions prone to misclassification and achieves superior performance on the challenging BraTS2021 dataset. These developments suggest that combining global contextual modeling with uncertainty awareness can significantly improve the robustness of brain tumor segmentation, particularly in handling ambiguous boundaries. However, further exploration is needed to effectively integrate these strategies with multimodal inputs and adaptive attention mechanisms—especially when dealing with complex tumor heterogeneity and modality-specific variations.

While recent advancements in brain tumor segmentation have demonstrated improved performance, there are limitations that hinder their efficacy in real-world clinical settings. Many current approaches struggle to simultaneously capture fine-grained local features and global contextual information, especially in the presence of complex tumor morphology and multimodal heterogeneity. Furthermore, models based on Transformers or uncertainty-guided attention mechanisms often suffer from high computational costs, increased model complexity, and suboptimal parameter efficiency. To address these challenges, we propose UPMAD-Net, an **U**ncertainty-guided **P**rior-integrated **M**ultiscale **A**ttention-based **D**ecoder Network. Our method integrates prior knowledge, multi-scale feature fusion, adaptive attention mechanisms, and probabilistic uncertainty estimation to achieve accurate and robust segmentation under complex conditions. UPMAD-Net is specifically designed to enhance segmentation precision in regions with ambiguous boundaries and structural variability, while maintaining low model complexity and parameter efficiency, making it more suitable for practical clinical applications.

The following are the main contributions of the paper:

(1) Prior Knowledge Integration: We incorporate prior knowledge related to tumor geometry, anatomical location, and appearance characteristics to guide model training, strengthen feature representation, and enhance segmentation robustness under complex conditions.

(2) Multi-Scale Feature Fusion: A multi-scale feature fusion module is embedded in each encoder layer, utilizing convolutional kernels of various sizes to expand the receptive field and improve the extraction of local and multi-scale semantic features.

(3) Adaptive Attention: mechanisms: An adaptive attention module is integrated into each decoder layer to dynamically reweight voxel-level features through attention mechanisms, enabling the model to focus more effectively on tumor regions and boundaries.

(4) Uncertainty Estimation: Monte Carlo Dropout (MC Dropout )is adopted during inference to estimate prediction uncertainty, thereby improving the reliability and interpretability of segmentation results, especially in ambiguous regions.

## 2. Method
### 2.1 Overview of UPMAD-Net

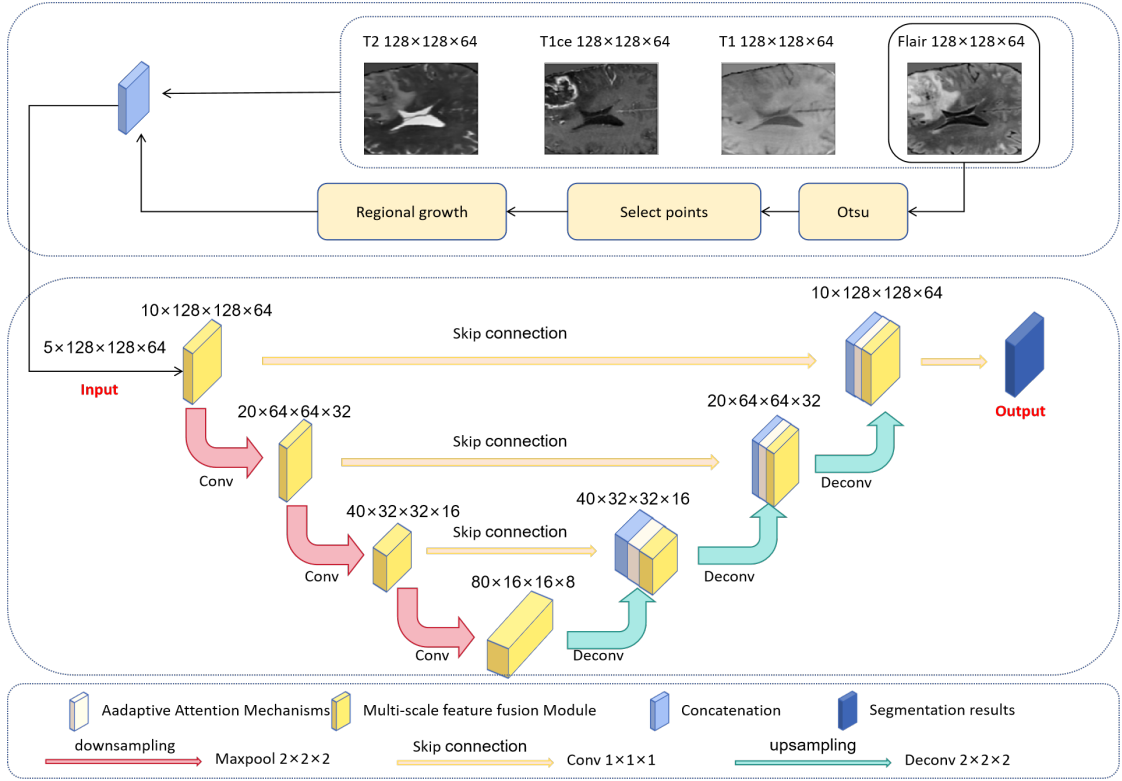

Fig. 1 Framework of the proposed UPMAD-Net.

As illustrated in Fig. 1, the proposed segmentation network incorporates a prior generation module that leverages traditional image processing techniques to provide structural guidance. Specifically, Otsu thresholding is first applied exclusively to the FLAIR modality to generate a coarse binary mask. The largest connected component is then extracted, and a certain number of seed points are randomly selected within this region to ensure that they are located in the most significant tumor area. Followed by 3D region growing to generate an initial tumor region. The resulting prior segmentation is concatenated with the four original MRI modalities, forming a five-channel input to the main network.

The core network adopts an encoder-decoder architecture inspired by U-Net, with several adaptive enhancements introduced at key stages. The encoder consists of four stages, each integrated with a multi-scale feature fusion(MSFF) module. MSFF modules apply parallel convolutional branches with varying kernel sizes to extract semantic features at multiple scales, enhancing the network's representational capacity. Downsampling is performed using 2×2×2 3D max pooling (MaxPool3D), effectively reducing spatial dimensions while preserving essential contextual information.

Unlike the symmetric structure of conventional U-Net, the decoder comprises three stages. Each decoding stage incorporates an adaptive attention mechanisms(AAM) and an MSFF module to progressively refine the upsampled features, allowing the model to focus more effectively on ambiguous boundaries and heterogeneous regions.

Upsampling is implemented using 2×2×2 3D transposed convolution (ConvTranspose3D), enabling learnable spatial reconstruction and precise feature alignment.

To facilitate efficient information flow and mitigate gradient vanishing, skip connections are introduced between the encoder and decoder. Each skip pathway includes a 1×1×1 convolution to adaptively recalibrate shallow encoder features, ensuring effective integration with deeper semantic representations and enhancing the model's ability to capture fine tumor boundaries and structural details.

## 2.2 Prior Module

To improve the model initial localization capability for tumor regions, pre-segmentation processing is conducted prior to model training. Specifically, an initial segmentation mask is generated using Otsu thresholding combined with region growing algorithms, serving as supplementary input during training.

The proposed segmentation framework introduces a prior generation module that utilizes traditional image processing techniques to provide structural guidance. Specifically, Otsu thresholding method is applied to the FLAIR modality to segment regions with intensities higher than the automatically determined threshold, which are assumed to be candidate regions for tumor localization. The largest connected component, defined as the region containing the greatest number of contiguous voxels, is then extracted. A subset of seed points is randomly selected within this component to initialize the region-growing process for subsequent tumor delineation. During the region-growing process, a grayscale difference threshold, defined as the median of the standard deviations of voxel intensities within annotated tumor regions across the training set, is used to constrain the inclusion of neighboring voxels. The refined prior segmentation map is then concatenated with the original four MRI modalities (T1, T1ce, T2, FLAIR), forming a five-channel input with both multimodal imaging features and anatomically constrained structural priors.

## 2.3 Multi-scale Feature Fusion Module

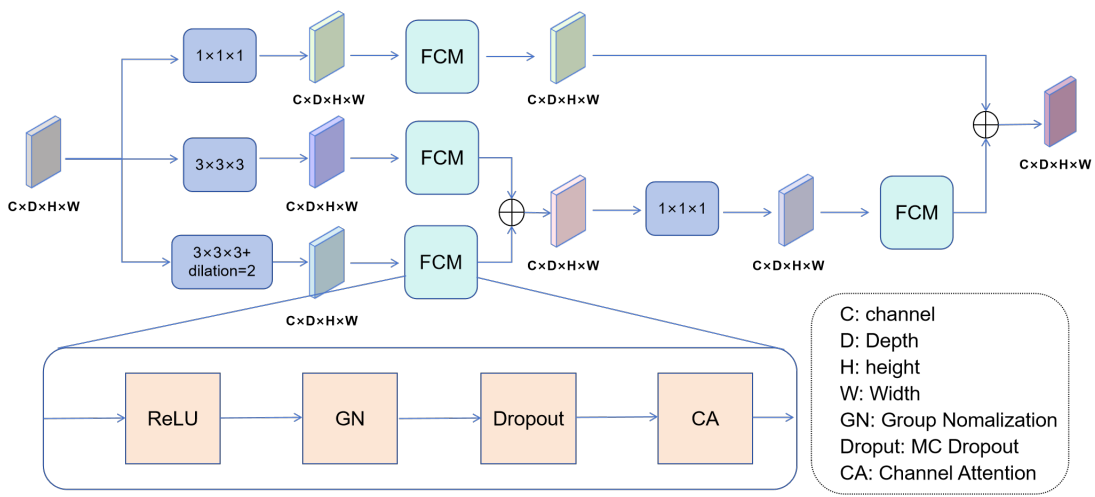

Fig. 2 Multi-level Feature Extraction Module.

As illustrated in Fig. 2, the MSFF enhances segmentation performance by enlarging the receptive field and reducing the loss of local information. The employment of multiple convolutional strategies enables the module to capture multi-scale features that preserve fine details while effectively modeling global context. This also improves the model's perceptual capability.

The input features are simultaneously processed through a 1×1×1 convolution, a standard 3×3×3 convolution, and a dilated 3×3×3 convolution with a dilation rate of 2 to extract tumor boundaries and texture details. The resulting features are then passed through the Feature Calibration Module (FCM), which includes ReLU activation to improve nonlinear representation, Group Normalization (GN) for training stability, MC Dropout for regularization, and a Channel Attention (CA) mechanism to highlight informative channels. Importantly, the use of dilated convolutions achieves a receptive field comparable to large-kernel convolutions without increasing the number of parameters or computational cost, allowing for efficient cross-region context modeling with minimal redundancy.

The outputs from the 3×3×3 and dilated convolutions are fused via element-wise addition and further refined through a 1×1×1 convolution. Finally, the output of the 1×1×1 convolution is added to the fused features through a residual connection, yielding the final integrated feature representation with enhanced continuity and diversity.

**2.4. Adaptive Attention Module**

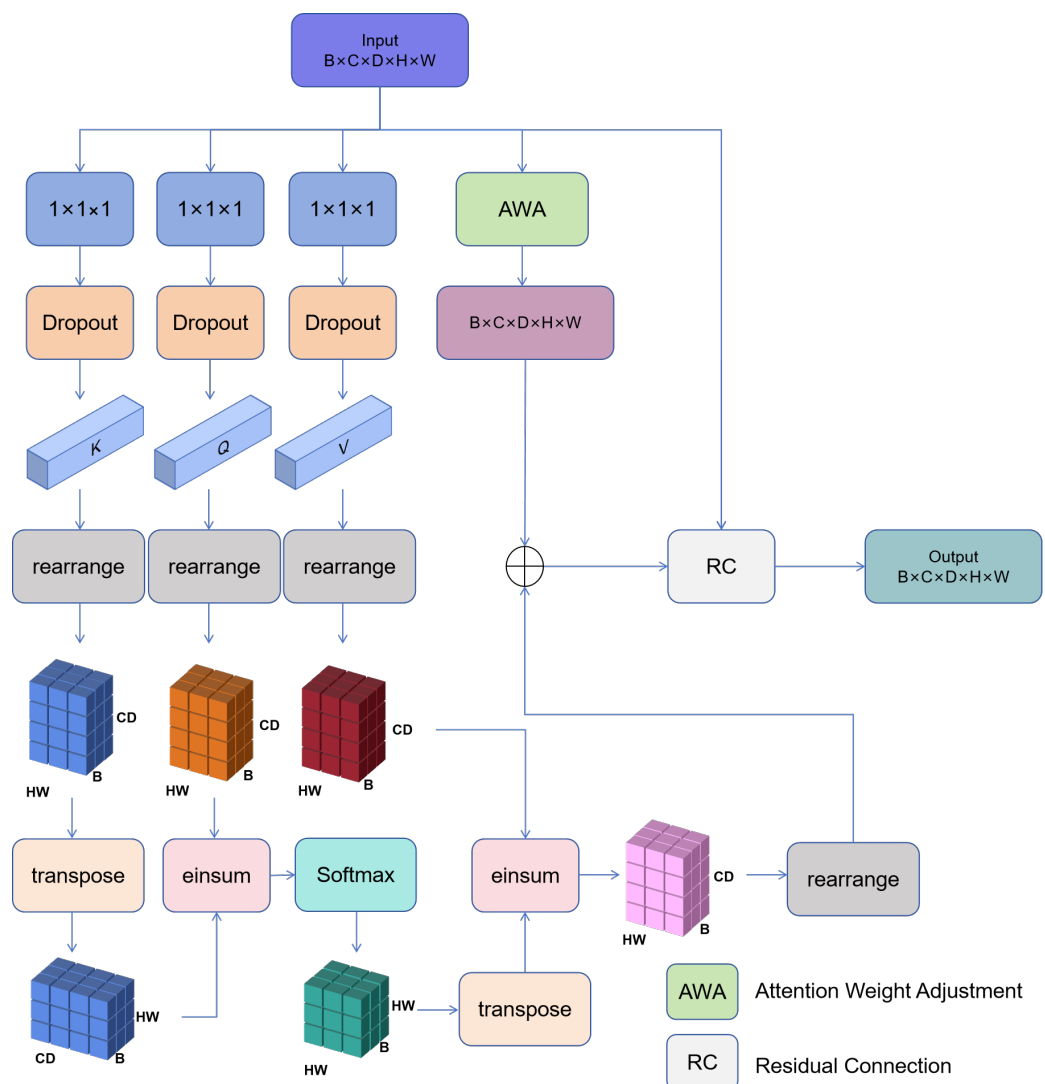

Fig. 3 Adaptive Attention Module.

Illustrated in Fig. 3, the AAM dynamically adjusts voxel-level attention weights during decoding to emphasize tumor regions and suppress background noise, thereby improving segmentation accuracy and robustness. The module takes an input tensor of shape B×C×D×H×W, where B is the batch size, C the number of channels, and D×H×W represents the spatial dimensions of the input volume.

The input is first processed through three parallel branches, each consisting of a 1×1×1 convolution followed by a Dropout layer, generating three feature maps denoted as K, Q, and V. These feature maps are reshaped into 3D tensors to enable efficient computation. The attention map is computed by performing an einsum operation between the transposed K and Q, followed by a Softmax function to obtain normalized attention scores. These scores are then combined with V via another einsum operation to produce an intermediate attention-enhanced output. Simultaneously, the input is fed into two additional pathways. One passes through an Attention Weight Adjustment (AWA) module, where the feature tensor is modulated element-wise with the intermediate output. The other follows a Residual Connection (RC) pathway. The outputs of these two paths are then merged and reshaped into the final feature representation. Lastly, a second residual connection integrates this output with the original input to produce the final output of the AAM.

**2.5 Training and Inference**

**Training:** The loss function utilized in this study consists of a combination of Dice loss ($L_{\text{DICE}}$) and binary cross-entropy loss ($L_{\text{BCE}}$). Dice loss measures the spatial overlap between the predicted and ground-truth regions, making it effective in capturing the structural characteristics of the segmented area. In contrast, BCE loss focuses on voxel-wise probability predictions, which helps address class imbalance but does not explicitly account for structural consistency. By combining these two loss functions, the model achieves a balance between spatial accuracy and voxel-level prediction reliability. The final combined loss function can be expressed mathematically as follows.

$$Loss = \frac{L_{\text{DICE}} + L_{\text{BCE}}}{2} \tag{1}$$

$$L_{\text{DICE}} = 1 - \frac{2\sum_{i=1}^{N}(p_i \times g_i) + \epsilon}{\sum_{i=1}^{N} p_i + \sum_{i=1}^{N} g_i + \epsilon} \tag{2}$$

Where $p_i$ represents the predicted value for the $i$-th pixel, $g_i$ represents the ground truth value for the $i$-th pixel, $N$ is the total number of pixels, and $\varepsilon$ is a small scalar to avoid division by zero.

$$L_{\text{BCE}} = -\frac{1}{N}\sum_{i=1}^{N}(g_i \log(p_i) + (1 - g_i)\log(1 - p_i)) \tag{3}$$

Where $p_i$ represents the predicted probability for the $i$-th pixel, $g_i$ denotes the ground truth label for the $i$-th pixel, $N$ is the total number of pixels, and denotes the natural logarithm.

**Inference:** Due to the complex and heterogeneous morphology of brain tumors, segmentation confidence often varies considerably across different regions. To enhance model robustness, especially in areas with low confidence, this study incorporates an uncertainty estimation strategy based on MC Dropout[23]. During the inference process, Dropout layers remain active, and multiple stochastic forward passes (e.g., 20 iterations) are performed to generate a series of predictions. The final segmentation result is obtained by averaging these predictions, while voxel-wise uncertainty is quantified by calculating the variance across the multiple outputs. This process yields uncertainty maps that highlight low-confidence regions, particularly those with ambiguous or indistinct tumor boundaries. As demonstrated in the uncertainty heatmaps presented in Section 4.5, MC Dropout effectively captures predictive uncertainty, providing intuitive visualizations that aid in identifying unreliable regions, guiding model refinement, and reducing over-reliance on any single prediction.

**3 Experiments**

**3.1 Dataset**

This study evaluates the proposed model on the BraTS2021 (1,251 cases) and BraTS2019 (335 cases) datasets, both of which share a consistent format. Each case includes four MRI modalities (FLAIR, T1ce, T1, and T2) and corresponding labels for background, necrosis, edema, and enhancing tumor regions. The segmentation targets include Enhancing Tumor (ET), Whole Tumor (WT), and Tumor Core (TC). Each MRI scan has a spatial resolution of 240×240×150 with a single-channel input. Since some slices contain no tumor regions, all scans were centrally cropped to 128×128×64 to accelerate training. While minor tumor areas may have been excluded, the cropping effectively removed irrelevant background and retained most tumor regions. As only training data is labeled in BraTS, we split the original training set into training, validation, and test subsets using an 8:1:1 ratio[24].

**3.2 Implementation Details**

The proposed UPMAD-Net is implemented using PyTorch[25]. The prior module takes a single-channel FLAIR image as input with dimensions 1×128×128×64, generating a rough prior segmentation. The network input is formed by concatenating four MRI modalities (FLAIR, T1ce, T1, and T2) with this prior map, resulting in a 5-channel volume of size 5×128×128×64. The batch size is set to 1 to accommodate the high memory demands of 3D multi-modal inputs and to preserve anatomical context without inter-sample interference. Training is conducted on a Tesla V100 GPU (32GB) for up to 1,000 epochs, using early stopping with a patience of 150 epochs to prevent overfitting. The model is optimized using the AdamW optimizer with an initial learning rate of 1e-4, weight decay of 1e-5, and a CosineAnnealingLR learning rate scheduler with epochs set to 50 epochs.

**3.3 Evaluation Metrics**

To quantitatively evaluate segmentation performance, this study employs two widely used metrics: Dice Similarity Coefficient ($Dice$) and Hausdorff Distance ($HD$). $Dice$ measures the degree of overlap between the predicted segmentation and the ground truth, directly reflecting the overall segmentation accuracy. In contrast, $HD$ evaluates the maximum distance between the boundaries of the predicted segmentation and the ground truth, thereby emphasizing boundary-level precision. In the context of medical image segmentation, these two complementary metrics provide a comprehensive assessment of both regional similarity and boundary alignment.

$$\text{Dice}(A, B) = \frac{2|A \cap B|}{|A| + |B|} \tag{4}$$

Where $A$ and $B$ represent the sets of voxels in the predicted and ground truth segmentations,

$$\text{Hausdorff}(P, G) = \max\left(\max_{p \in P} \min_{g \in G} |p - g|, \max_{g \in G} \min_{p \in P} |g - p|\right) \tag{5}$$

Where $P$ and $G$ denote the sets of boundary points. The Euclidean distance between two points is used for the $HD$ calculation.

**4 Experimental Results**

**4.1 Results of segmentation**

To comprehensively evaluate the effectiveness of UPMAD-Net in brain tumor segmentation, we conducted experiments on the BraTS2019 and BraTS2021 datasets, using Dice and HD metrics to assess three tumor sub-regions: ET, WT, and TC. The performance of our method was compared against several state-of-the-art approaches, with all results sourced directly from their respective publications for fair comparison.

As shown in Table 1, Table 2, and Table 3, UPMAD-Net achieves superior or highly competitive performance across all datasets and metrics. On the BraTS2021 validation set, UPMAD-Net obtains Dice scores of 90.00% (ET), 94.72% (WT), and 92.42% (TC), with corresponding HD values of 2.450 mm, 2.882 mm, and 2.381 mm, surpassing MPEDA-Net, the second-best in most sub-regions. On the test set, it maintains high performance, with Dice scores of 89.18% (ET), 93.67% (WT), and 91.23% (TC), and HD values of 2.420 mm, 3.020 mm, and 2.461 mm, respectively.

On the BraTS2019 validation set, UPMAD-Net achieves Dice scores of 87.33% (ET), 90.81% (WT), and 90.22% (TC), with HD values of 2.557 mm, 2.376 mm, and 2.607 mm, outperforming Residual U-Net across all sub-regions. For instance, while Residual U-Net achieves a comparable Dice score for WT (88.66% vs. 90.81%), its HD is substantially worse (10.340 mm vs. 2.376 mm), indicating better boundary localization by UPMAD-Net. On the BraTS2019 test set, UPMAD-Net continues to demonstrate strong generalization ability, achieving Dice scores of 84.03% (ET), 91.68% (WT), and 86.89% (TC), with corresponding HD values of 2.720 mm, 3.351 mm, and 2.947 mm.

The qualitative results in Fig. 4 visually confirm these findings. Red contours denote the predictions from UPMAD-Net, and yellow contours indicate the ground truth. Across both BraTS2019 and BraTS2021 datasets, the predictions tightly align with the ground truth, even in complex and subtle tumor cases, highlighting the model's superior boundary delineation and generalization ability.

UPMAD-Net demonstrates better performance on BraTS2021 than BraTS2019, likely due to improved image quality, more consistent annotations, and advanced preprocessing protocols. Despite domain discrepancies, UPMAD-Net exhibits robust cross-dataset generalization, attributed to its multiscale contextual modeling and effective integration of prior knowledge.

**Table 1:** Comparison of the segmentation performance on the BraTS2021 validation dataset. The bold and underlined indicate the best and second best.

| Method | Dice(%) | | | HD(mm) | | |
|---|---|---|---|---|---|---|
| | ET | WT | TC | ET | WT | TC |
| AMAF-Net[26] | 78.65 | 90.45 | 82.13 | 4.441 | 7.052 | 4.999 |
| ASOU-Net[27] | 77.67 | 86.74 | 80.19 | 5.053 | 6.366 | 6.460 |
| HDC-Net[28] | 72.48 | 75.74 | 75.57 | 7.606 | 7.144 | 7.564 |
| BraTS-DMFNet[29] | 75.99 | 86.73 | 80.66 | 7.995 | 7.969 | 7.990 |
| RALU-Net[30] | 82.31 | 91.97 | 86.84 | 2.421 | 3.903 | 2.815 |
| Att_EquiUnet[31] | 72.07 | 91.12 | 86.61 | 2.325 | 3.324 | 2.584 |
| IAU-Net[32] | 81.62 | 91.60 | 86.12 | 2.282 | 3.524 | 2.482 |
| Swin-UnetR[33] | 85.80 | 92.60 | 88.50 | 6.016 | 5.831 | 3.770 |
| MPEDA-Net[34] | 82.52 | 93.07 | 87.67 | **2.204** | 3.219 | **2.318** |
| Ours | **90.00** | **94.72** | **92.42** | 2.450 | **2.882** | 2.381 |

**Table 2:** Comparing results with the other methods on the BraTS2021 test dataset.

| Method | Dice(%) | | | HD(mm) | | |
|---|---|---|---|---|---|---|
| | ET | WT | TC | ET | WT | TC |

| Method | | | | | | |
|---|---|---|---|---|---|---|
| AMAF-Net | 77.46 | 89.94 | 81.09 | 4.470 | 6.980 | 5.021 |
| ASOU-Net | 77.31 | 87.00 | 78.79 | 5.137 | 6.335 | 6.462 |
| HDC-Net | 72.01 | 77.00 | 74.26 | 7.578 | 7.108 | 7.527 |
| BraTS-DMFNet | 75.45 | 87.77 | 79.50 | 7.993 | 7.961 | 7.986 |
| RALU-Net | 80.32 | 91.12 | 85.01 | 2.468 | 3.935 | 2.817 |
| Att_EquiUnet | 80.42 | 90.89 | 84.64 | 2.390 | 3.383 | 2.655 |
| IAU-Net | 79.82 | 91.27 | 82.98 | 2.341 | 3.602 | 2.619 |
| Swin-UnetR | 85.30 | 92.70 | 87.60 | 16.326 | 4.739 | 15.309 |
| MPEDA-Net | 81.14 | 92.12 | 84.79 | **2.273** | 3.279 | **2.414** |
| Ours | **89.18** | **93.67** | **91.23** | 2.420 | **3.020** | 2.461 |

**Table 3:** Comparing results with the other methods on the BraTS2019 validation dataset.

| Method | Dice(%) | | | HD(mm) | | |
|---|---|---|---|---|---|---|
| | ET | WT | TC | ET | WT | TC |
| SoResU-Net[35] | 72.40 | 87.50 | 78.80 | 5.970 | 9.350 | 11.470 |
| KiU-Net[36] | 73.20 | 87.60 | 73.90 | 6.320 | 8.940 | 9.890 |
| LCRLNet[37] | 72.70 | 87.10 | 71.80 | 6.300 | 6.700 | 9.300 |
| AMMGS[38] | 76.80 | 89.3 | 81.10 | 5.180 | 8.220 | 7.230 |
| TransBTS[39] | 73.70 | 89.40 | 80.70 | 5.990 | 5.680 | 7.360 |
| AugTransU-Net[40] | 78.20 | 89.70 | 80.40 | 4.110 | 6.650 | 9.560 |
| Attention U-Net[41] | 69.36 | 87.48 | 76.26 | 12.260 | 17.790 | 16.74 |
| Residual U-Net[41] | 72.96 | 88.66 | 78.97 | 8.070 | 10.340 | 10.50 |
| Ours | **87.33** | **90.81** | **90.22** | **2.557** | **2.376** | **2.607** |

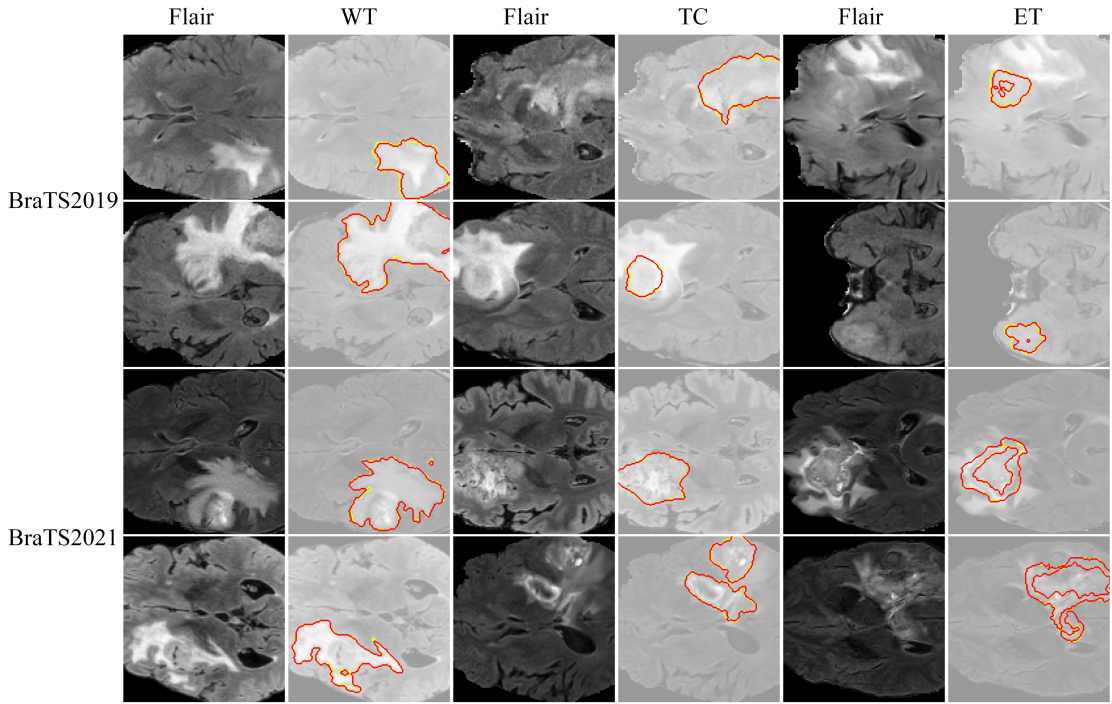

Fig. 4 Visualization of comparing segmentation results.

### 4.2 Ablation experiments

In the experiments, a modified U-Net with four convolutional and three pooling layers was used as the baseline to ensure depth alignment and skip connections with the proposed model. All models in Table 4 were trained for 50 epochs, based on convergence observed during preliminary experiments. Results were validated on the BraTS 2019 validation set.Adding only the Prior module yielded limited improvement over the baseline, suggesting that prior knowledge alone has marginal impact. In contrast, integrating the MSFF module with MC (MC Dropout ) significantly enhanced both Dice and HD scores across ET, WT, and TC regions, demonstrating its effectiveness in multi-scale feature fusion and boundary extraction. Although the AAM module improved voxel-level accuracy, it was less effective in boundary delineation compared to MSFF. Combining AAM, MSFF, and MC led to further improvements, indicating their complementarity—AAM refines voxel prediction, MSFF strengthens boundaries, and MC improves model robustness. UPMAD-Net, integrating all modules, achieved the best performance across all metrics.

**Table 4:** Results of Module Comparison Experiments.

| Method | Module | | | | Dice(%) | | | HD(mm) | | |
|---|---|---|---|---|---|---|---|---|---|---|
| | Prior | MSFF | AAM | MC | ET | WT | TC | ET | WT | TC |
| Baseline | | | | | 74.84 | 61.54 | 65.70 | 2.950 | 4.511 | 3.481 |
| Prior | √ | | | | 73.45 | 67.79 | 62.68 | 3.111 | 4.190 | 3.746 |
| MSFF+ MC | | √ | | √ | 81.36 | 87.54 | 81.39 | 2.807 | 3.631 | 3.115 |
| AAM+ MC | | | √ | √ | 73.05 | 80.07 | 60.02 | 3.320 | 3.991 | 3.898 |
| MSFF+AAM+MC | | √ | √ | √ | 82.84 | 88.69 | **83.86** | **2.808** | **3.488** | 3.059 |
| Full Model | √ | √ | √ | √ | **83.17** | **88.92** | 82.92 | 2.840 | 3.491 | **2.976** |

Fig. 5 visually compares segmentation results for a BraTS 2019 sample under six model variants. Red contours indicate ground truth; yellow contours show predictions. From Model 1 to Model 6, segmentation accuracy improves progressively. Earlier models suffer from over-segmentation or boundary mismatch, while later models, especially Model 6 (UPMAD-Net), produce accurate and compact segmentations aligned with the ground truth, confirming the method's effectiveness and robustness.

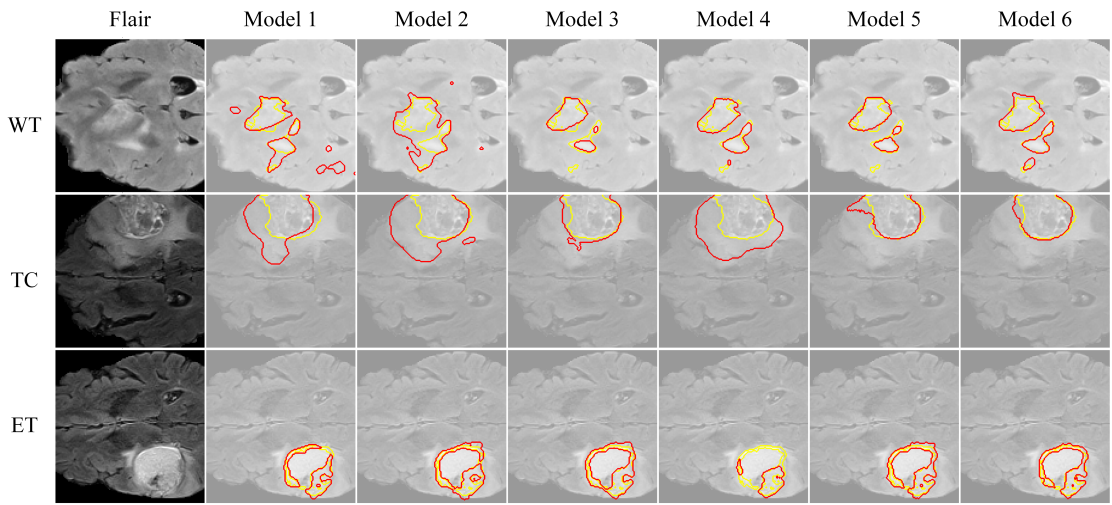

Fig. 5 Visual comparison of different model combinations on the BraTS2019 validation set .Ground truth annotations are outlined in red, while predicted segmentation results are shown in yellow.(Model 1:

Baseline; Model 2: Baseline + Prior; Model 3: Baseline + MSFF + MC; Model 4: Baseline + AAM + MC; Model 5: Baseline + AAM + MSFF + MC; Model 6: UPMAD-Net).

### 4.3 Efficiency Analysis of Kernel Size and Dilation in the MSFF Module

To intuitively illustrate the effectiveness of integrating large convolutional kernels with dilation rates in the MSFF module for reducing computational and memory costs, two sets of comparative experiments were conducted using the BraTS2019 validation dataset Table 5.

presents a comparative evaluation of multiple methods, including UPMAD-Net, in terms of floating point operations per second (FLOPs), number of trainable parameters (Params), and inference time. As illustrated, UPMAD-Net achieves one of the lowest FLOPs (20.01G) and parameter counts (0.34M) among the compared methods. Although its inference time of 74.0 ms is not the shortest, UPMAD-Net still exhibits competitive performance while maintaining high segmentation accuracy.

Table 6 illustrates the impact of various kernel sizes and dilation rates on UPMAD-Net segmentation performance. Specifically, two configuration types were compared: (1) large convolutional kernels ($5\times5\times5$ and $7\times7\times7$) and (2) $3\times3\times3$ kernels with varying dilation rates (2 and 3). Experimental results indicate that while larger kernels can slightly improve segmentation performance, they substantially increase computational cost, inference time, and parameter count. In contrast, applying an appropriate dilation rate (2 or 3) to $3\times3\times3$ kernels yields only 20.01G FLOPs and 0.34M parameters, faster inference, and comparable segmentation accuracy for ET, WT, and TC.These findings suggest that in practical applications, adopting smaller kernels with suitable dilation is a more efficient network design strategy. This approach not only decreases computational and memory overhead and accelerates inference but also reduces model complexity while maintaining or even enhancing segmentation accuracy. Compared with directly employing large kernels (e.g. $5\times5\times5$ or $7\times7\times7$), this configuration is evidently more advantageous for practical deployment.

**Table 5:** Comparision of FLOPs, parameters, and inference time.

| Method | FLOPs | Param. | Inference time |
|---|---|---|---|
| AMAF-Net | 20.34G | 1.94M | **0.0307s** |
| ASOU-Net | 88.12G | 6.48M | 0.681s |
| HDC-Net | 183.24G | 0.55M | 0.1523s |
| BraTS-DMFNet | 13.20GG | 3.88M | 0.0616s |
| RALU-Net | 174.05G | 8.52M | 0.1021s |
| IAU-Net | 66.34G | 16.61M | 0.0938s |
| MPEDA-Net | 49.68G | 0.87M | 0.1854s |
| UPMAD-Net(Ours) | **20.01G** | **0.34M** | 0.1717s |

**Table 6**: Comparison of Different size of Convolution Kernel in UPMAD-Net.

| Method | Flops | Param | Inference times | Dice(%) | | |
|---|---|---|---|---|---|---|
| | | | | ET | WT | TC |
| UPMAD-Net(5x5x5) | 49.99G | 0.87M | 0.1932s | **83.74** | 87.80 | **83.97** |
| UPMAD-Ne(7x7x7) | 116.74G | 2.05M | 0.3862s | 82.00 | 87.84 | 83.17 |
| UPMAD-Ne(3x3x3+3) | **20.01G** | **0.34M** | 0.1763s | 82.10 | 88.83 | 82.23 |
| UPMAD-Net(3x3x3+2) | **20.01G** | **0.34M** | **0.1717s** | 83.17 | **88.92** | 82.92 |

### 4.4 Selection of the Gray-Level Difference Threshold

To determine a suitable gray-level difference threshold for seed-point-based region growing, we first computed the standard deviations of voxel intensities within tumor regions across the training set. As shown in Fig. 6, the standard deviations range from 15.43 to 71.53, with a median of 33.71 and most values falling between 20 and 50. Based on this distribution, we selected a threshold of 35 to account for intra-tumoral heterogeneity. This value was chosen for two main reasons: (1) it closely aligns with the median standard deviation, ensuring broad coverage of typical cases; and (2) it strikes a balance between avoiding under-segmentation from overly low thresholds and over-segmentation from excessively high ones. Thus, setting the threshold at 35 enables more accurate and robust tumor boundary delineation across varied lesion types.

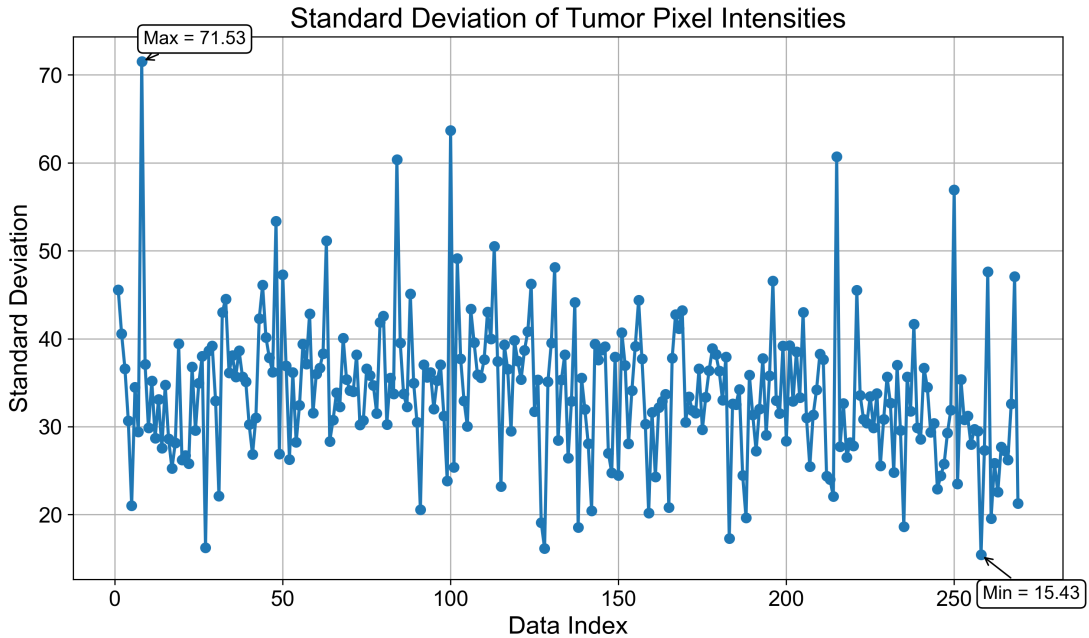

Fig. 6 Distribution of Standard Deviations Across Tumor Regions (Min: 15.43, Max: 71.53, Median: 33.71).

### 4.5 Analysis of Uncertainty-Guided Boundary Localization

To validate the effectiveness of incorporating uncertainty estimation (via MC Dropout) in brain tumor segmentation, a visual analysis was conducted using uncertainty heatmaps on the test dataset. As shown in Fig. 7, four key views are presented: the original FLAIR image, predicted vs. ground truth contours, prediction probability maps, and uncertainty heatmaps.

The original FLAIR image provides structural context, allowing for a better understanding of tumor morphology and localization. The comparison between the model's predictions and the ground truth reveals that the model captures most tumor regions accurately, with red contours indicating predicted boundaries and yellow contours denoting the ground truth; minor deviations are primarily observed along tumor edges. The prediction probability map illustrates the model's confidence levels, where darker red shades correspond to higher certainty, and reduced confidence is generally observed near the tumor periphery, indicating areas of increased ambiguity. Correspondingly, the uncertainty heatmap, ranging from black (low uncertainty) to white (high uncertainty), highlights these boundary regions with elevated uncertainty, reflecting the model's reduced confidence in classifying such ambiguous areas.

These results demonstrate that integrating uncertainty quantification (UQ) enables the model to identify and localize ambiguous tumor boundaries and potential error regions. This provides clinicians with an intuitive understanding of prediction confidence and supports further investigation or additional imaging in high-uncertainty areas, thereby improving diagnostic accuracy. Furthermore, during model development, UQ helps detect dataset or architectural issues, guiding targeted improvements and data augmentation strategies to enhance overall model robustness and reliability.

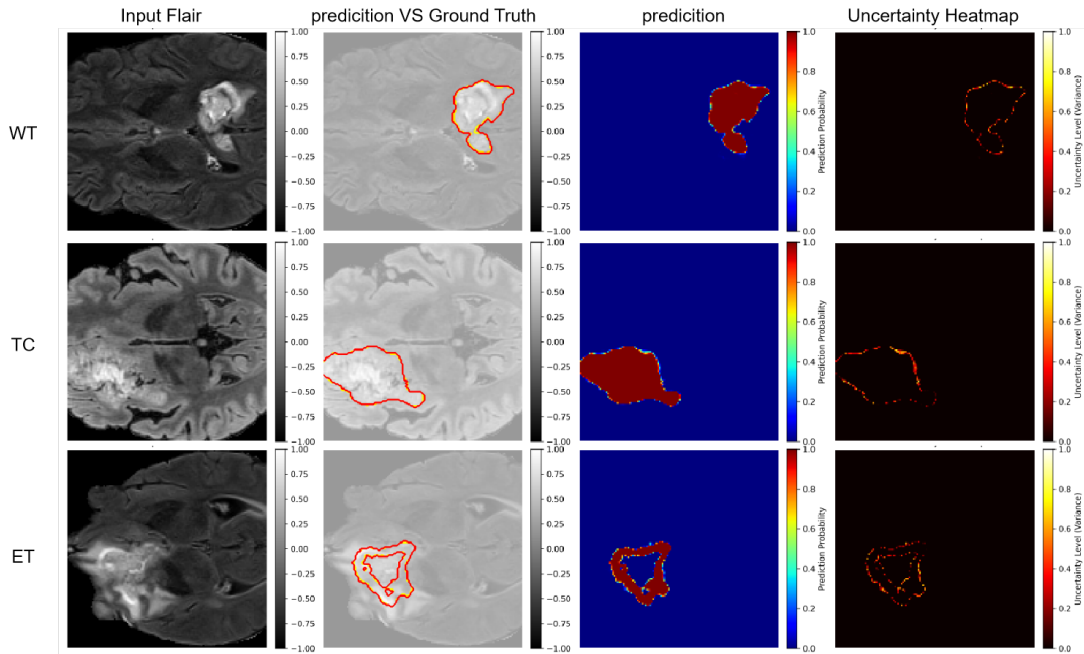

Fig. 7 Multislice Visualization of Prediction and Uncertainty.

## 5 Conclusion

This study proposed a novel 3D brain tumor segmentation network based on the U-Net architecture. An MSSF module was embedded in the encoder to expand the receptive field while preserving fine-grained spatial details using diverse convolutional kernels. An AAM was incorporated into the decoder to enhance multi-level contextual integration and ensure smooth feature transitions. Prior knowledge provided reliable localization cues, improving segmentation accuracy and efficiency. MC Dropout was used to estimate prediction uncertainty, enabling the identification of low-confidence regions and enhancing model robustness in ambiguous areas.

Future work will explore advanced attention mechanisms and Transformer-based architectures to improve contextual modeling and maintain stability. Additionally, we plan to design more efficient 3D convolutional modules to strengthen the network's ability to capture local structural information.

**Declaration of competing interest:** The authors declare that they have no known competing financial interests or personal relationships that could have appeared to influence the work reported in this paper.

**Acknowledgments:** This study received support from the National Natural Science Foundation of China (Grant Numbers: 62476255, 62303427, 82370513 and 62106233), the Science and Technology Innovation Talent Project of Henan Province University under Grant 25HASTIT028, and the Henan Science and Technology Development Plan (Grant Number: 232102210010, 232102210062). Zhongyuan Science and Technology Innovation Outstanding Young Talents Program.